\documentclass[10pt,twocolumn,letterpaper,table]{article}

\usepackage{iccv}              

\definecolor{iccvblue}{rgb}{0.21,0.49,0.74}
\usepackage[pagebackref,breaklinks,colorlinks,allcolors=iccvblue]{hyperref}

\usepackage{color}
\usepackage{soul}
\usepackage{graphicx}

\usepackage{booktabs}
\usepackage{booktabs}
\definecolor{good}{HTML}{A6D785}
\definecolor{bad}{HTML}{F28E8E}

\def\paperID{*****} 
\def\confName{ICCV}
\def\confYear{2025}

\def\ours{DiViD}

\newcommand\nframes{N}

\title{DiViD: Disentangled Video Diffusion for Static–Dynamic Factorization}

\author{
	Marzieh Gheisari \quad Auguste Genovesio \\
	École Normale Supérieure PSL, Paris, France \\
	{\tt\small \{marzieh.gheisari, auguste.genovesio\}@ens.psl.eu}
}

\begin{document}
\pagestyle{empty}
\thispagestyle{empty}
\maketitle

\begin{abstract}
Unsupervised disentanglement of static appearance and dynamic motion in video remains a fundamental challenge, often hindered by information leakage and blurry reconstructions in existing VAE- and GAN-based approaches. We introduce \textbf{DiViD}, the first end-to-end \emph{video diffusion} framework for explicit static–dynamic factorization. DiViD’s sequence encoder extracts a global \emph{static token} from the first frame and per-frame \emph{dynamic tokens}, explicitly removing static content from the motion code. Its conditional DDPM decoder incorporates three key inductive biases: a \emph{shared-noise schedule} for temporal consistency, a \emph{time-varying KL-based bottleneck} that tightens at early timesteps (compressing static information) and relaxes later (enriching dynamics), and \emph{cross-attention} that routes the global static token to all frames while keeping dynamic tokens frame-specific. An orthogonality regularizer further prevents residual static–dynamic leakage.

We evaluate DiViD on real-world benchmarks using swap-based accuracy and cross-leakage metrics. DiViD outperforms state-of-the-art sequential disentanglement methods: it achieves the highest swap-based joint accuracy, preserves static fidelity while improving dynamic transfer, and reduces average cross-leakage.
\end{abstract}

\section{Introduction}
\label{sec:intro}
The field of representation learning faces a fundamental challenge in unsupervised disentanglement, which aims to decompose input data into its latent factors of variation. This approach is crucial for improving machine learning tasks by enhancing \emph{explainability}, \emph{generalizability}, and \emph{controllability}~\cite{locatello2019disentangling,liu2022learning,wang2022disentangled,fragemann2022review,bengio2013representation}
In the context of sequential data such as video, the goal is specifically to separate latent representations into a \emph{single static (time‐invariant) factor} and \emph{multiple dynamic (time‐varying) components}. For example, in a video of a person smiling, the person’s identity is the static component, while the smiling motion is the dynamic component. Such disentangled representations are highly beneficial for downstream tasks including classification, prediction, retrieval, interpretability, and synthetic video generation with style transfer.
The present work, in particular, focuses on unsupervised sequential disentanglement\cite{villegas2017decomposing, tian2021good, tulyakov2018mocogan}, which—unlike static disentanglement~\cite{yang2023disdiff, kingma2013auto, wang2022disentangled,zhu2021commutative, locatello2020weakly, bouchacourt2018multi, mita2021identifiable, higgins2016beta}—must exploit the inherent temporal structure of video data to improve factor separation and temporal coherence.

Most prior methods for sequential disentanglement build on Variational Autoencoders (VAEs) and their dynamic extensions\cite{bai2021contrastively,berman2024sequential,naiman2023sample}. While VAEs are deep generative probabilistic models that can learn disentangled representations with appropriate regularization, they face several challenges in the sequential setting:
\begin{itemize}
	\item \textbf{Information Leakage.} Conditioning static and dynamic factors on the entire input sequence often allows dynamic codes to capture static information (and vice versa), resulting in poor disentanglement. Prior remedies—e.g.\ reducing the dynamic latent dimension or adding auxiliary mutual‐information losses—tend to compromise dynamic expressiveness, complicate training with multiple loss terms, and exhibit sensitivity to hyperparameters.
	\item \textbf{Reconstruction Quality.} VAEs frequently generate blurry outputs on complex real‐world data. Techniques to sharpen reconstructions typically introduce hierarchical latent spaces that can impede disentanglement.
	\item \textbf{Insufficiency of Regularization.} Empirical studies show that relying solely on generic regularizers is insufficient. Effective disentanglement often requires explicit \emph{inductive biases} in both model architectures and training procedures.
\end{itemize}

GAN‐based approaches have incorporated regularizations to encourage disentangled feature learning, but their disentanglement capabilities remain less than satisfactory, and unsupervised disentangled representation learning with GANs continues to be very challenging~\cite{zhu2021and,chen2016infogan}.
More recently, diffusion models have emerged as powerful generative models, demonstrating superior visual quality compared to both VAEs and GANs~\cite{yang2023diffusion}. However, early diffusion architectures lacked semantic structure in their latent variables, making them suboptimal for disentanglement. Diffusion Autoencoders (DiffAEs)~\cite{preechakul2022diffusion} began to address this by learning meaningful representations, but they were tailored to non‐sequential data and did not explicitly factorize into static and dynamic components. Other diffusion‐based video methods (e.g.\ DiffusionVideoAutoencoder~\cite{kim2023diffusion})often rely on pretrained encoders, are domain‐specific.

\bigskip
\noindent
{We introduce \emph{\ours~: Disentangled Video Diffusion for Static–Dynamic Factorization.} \ours~is the first diffusion‐based framework designed end‐to‐end for unsupervised sequential disentanglement. It incorporates several key innovations and complementary inductive biases that drive clean disentanglement and overcome the challenges faced by prior methods:

\begin{itemize}
	\item \textbf{Architectural Bias for Leakage Mitigation}: Inspired by the idea of first frame encoding and using residuals for dynamic encoding from DBSE~\cite{berman2024sequential}, our sequence encoder incorporates a {subtraction mechanism}. This design effectively removes static features from dynamic factors by subtracting the latent representation of the first frame from subsequent frames, compelling the dynamic encoder to focus exclusively on temporal variations and directly mitigating information leakage.
	
\item \textbf{Diffusion-driven Inductive Biases}: Building on EncDiff~\cite{yang2024diffusion}'s findings, \ours~leverages two powerful inherent inductive biases within diffusion models that are conducive to disentanglement:
	\begin{itemize}
		\item {Time-Varying Information Bottleneck}: Our diffusion process inherently imposes a Kullback-Leibler (KL)-based information bottleneck that varies with time. At \emph{early timesteps} (large $t$), this bottleneck forces the static token to carry only the most essential, time-invariant information. As the bottleneck gradually relaxes at \emph{later timesteps} ($t \to 0$), dynamic tokens are enabled to capture richer, more detailed, frame-specific features, thereby promoting effective disentanglement.
		\item {Cross-Attention Interaction}: EncDiff demonstrated that cross-attention within diffusion models serves as a strong inductive bias for disentanglement in image generation. \ours~integrates this cross-attention within its U-Net denoiser, where every cross-attention block attends jointly to the global static token and the per-frame dynamic token. This mechanism explicitly routes global static information to consistently influence all frames, while local dynamic information affects only its corresponding frames, effectively ensuring a clear separation between the static and dynamic components.
	\end{itemize}
	
	\item \textbf{Enhanced Temporal Consistency with Shared Noise}: \ours~further enhances temporal consistency in generated sequences by employing a shared noise $\varepsilon$ across all frames during the diffusion process. This design choice is expected to enable more consistent information encoding throughout the video.
	
	\item \textbf{End-to-End Training with Orthogonality Regularization}: \ours~is trained end-to-end using a straightforward DDPM loss augmented by an {orthogonality regularization term} to further prevent static information leakage into dynamic codes.
\end{itemize}

We evaluate \ours~on real-world datasets, including MHAD and MEAD, demonstrating superior performance in disentanglement quality compared to existing state-of-the-art methods. Our framework achieves strong informativeness while significantly reducing cross-leakage, proving its effectiveness in separating static appearance and dynamic motion.
\begin{figure*}[t]
	\centering
	\includegraphics[width=0.9\linewidth]{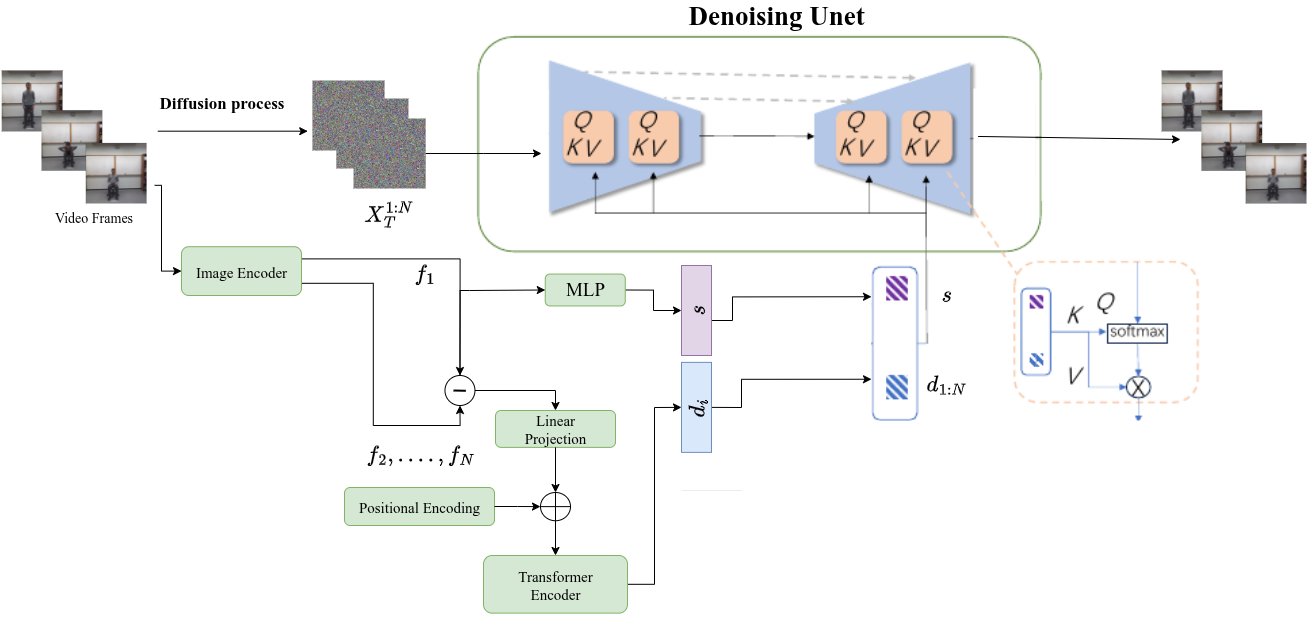}
	\caption{Overview of \ours. The sequence encoder decomposes an input video into a shared static token and frame-specific dynamic tokens, which condition the U-Net denoiser via cross-attention in the diffusion decoder.}
	\label{fig:framework}
\end{figure*}
\section{Related Work}
\label{sec:related}

\paragraph{Sequential Disentanglement}
Sequential disentanglement specifically addresses data with temporal dynamics, such as video sequences, aiming to separate latent factors into static (time-invariant) and dynamic (time-varying) components~\cite{villegas2017decomposing, tulyakov2018mocogan}. Early methods condition latent variables either on mean past features or directly on feature sequences~\cite{hsu2017unsupervised, li2018disentangled}. However, these approaches typically lead to information leakage, where static factors contaminate dynamic representations or vice versa. Remedies such as introducing auxiliary mutual information (MI) losses or reducing latent dimensions only achieve partial success due to sensitivity to hyperparameters and difficulty capturing complex dynamics~\cite{zhu2020s3vae, han2021disentangled}.

Contrastive learning methods, such as the Sample and Predict Your Latent (SPYL) approach, have addressed these issues by employing modality-free contrastive estimation strategies, enabling better disentanglement without relying on complex MI estimations or modality-specific augmentations~\cite{naiman2023sample, bai2021contrastively}. Furthermore, recent models like DBSE leverage the first-frame encoding and residual dynamics to explicitly isolate static and dynamic information, significantly mitigating leakage issues~\cite{berman2024sequential}.
\paragraph{Diffusion Models for Disentanglement}
Diffusion models have recently emerged as powerful generative models capable of surpassing VAEs and GANs in visual quality~\cite{ho2020denoising}. However, early diffusion architectures lacked structured semantic representations, limiting their applicability for disentanglement tasks. Diffusion Autoencoders (DiffAEs) introduced latent semantic structures within diffusion frameworks but primarily focused on non-sequential data without explicitly factorizing latent variables into static and dynamic components~\cite{kingma2021variational, preechakul2022diffusion}. Recent video-based diffusion models, such as DiffusionVideoAutoencoder, often rely on pretrained encoders and remain domain-specific, further limiting general applicability~\cite{kim2023diffusionvideoautoencoder}.

EncDiff demonstrated that diffusion models equipped with cross-attention mechanisms inherently provide powerful inductive biases for disentanglement, employing a time-varying information bottleneck and structured cross-attention~\cite{yang2024diffusion}. Building upon these insights, our proposed framework, DiViD, introduces a novel diffusion-based approach tailored explicitly for unsupervised sequential disentanglement. DiViD integrates architectural biases to mitigate information leakage by subtracting static features from dynamic encodings, leverages the intrinsic KL-based time-varying bottleneck characteristic of diffusion models, and employs cross-attention interactions within the diffusion denoiser to explicitly separate static and dynamic components. Furthermore, DiViD introduces shared noise across frames to enhance temporal coherence and employs orthogonality regularization to ensure robust disentanglement between static and dynamic representations.

\section{Method}
\label{sec:method}

We propose \ours, a novel diffusion-based framework specifically designed for unsupervised sequential disentanglement. \ours~integrates a sequence encoder that factorizes video sequences into distinct static and dynamic representations, combined with a conditional diffusion decoder to reconstruct video frames. We describe our overall framework in subsection~\ref{sec:framework}, highlight key inductive biases that encourage disentanglement in subsection~\ref{sec:biases}, and detail our end-to-end training approach in subsection~\ref{sec:training}.

\subsection{Framework}
\label{sec:framework}

Figure~\ref{fig:framework} illustrates the proposed architecture of \ours. Given a video sequence $v = \{x_{1},\dots,x_{\nframes}\}$, our sequence encoder $\tau_{\phi}$ generates a single static token $s$, representing time-invariant content, and dynamic tokens $\{d_{1},\dots,d_{\nframes}\}$, capturing frame-specific temporal information. For simplicity, we omit explicit indexing of video sequences in subsequent notation.

These static and dynamic tokens condition the diffusion decoder, implemented as a Denoising Diffusion Probabilistic Model (DDPM)~\cite{ho2020denoising}. In the forward diffusion process, each frame $x_i$ is corrupted with Gaussian noise $\epsilon \sim \mathcal{N}(0,I)$ at timestep $T$, yielding a noisy observation:
\begin{equation}
x_{T,i} = \sqrt{\bar\alpha_T}\,x_i + \sqrt{1 - \bar\alpha_T}\,\epsilon,
\end{equation}
where $\bar\alpha_T$ defines the noise scheduling. Importantly, the noise realization $\epsilon$ is shared across all frames, an intentional design choice aimed at improving temporal consistency.

The diffusion decoder reconstructs the frames through an iterative reverse process:
\begin{equation}
p_\theta(x_{i,0:T}\mid s, d_i) = p(x_{i,T})\prod_{t=1}^T p_\theta(x_{i,t-1}\mid x_{i,t}, s, d_i),
\end{equation}
conditioned explicitly on the static and dynamic tokens.

\subsection{Inductive Biases for Disentanglement}
\label{sec:biases}

To achieve effective static-dynamic disentanglement, \ours{} leverages three complementary inductive biases inspired by successful principles from recent work but introduces significant novel adaptations to enhance disentanglement performance:

\paragraph{1. Architectural Bias (Static–Dynamic Residual Encoding)}
Inspired by DBSE \cite{berman2024sequential}, we adopt first-frame encoding and frame-residual features. We extend this idea with a \emph{Transformer}-based sequence encoder—multi-head self-attention with feed-forward blocks, residual connections, and layer normalization. Residuals \(r_i = f_i - f_1\) are linearly projected to the model width \(z_{\mathrm{dim}}\), augmented with positional encodings \(\mathrm{PE}(i)\), and processed by a stack of Transformer encoder layers to yield frame-specific dynamic tokens \(d_i\). This residual formulation discourages leakage of static information into the dynamic representation.


\paragraph{2. Time-Varying Information Bottleneck}
Drawing on ideas from EncDiff \cite{yang2024diffusion}, our diffusion decoder leverages a timestep-dependent KL-based bottleneck. Specifically, at early diffusion timesteps (high $t$), a tight bottleneck constrains static tokens $s$ to encode only essential, time-invariant information. As $t$ decreases, the bottleneck relaxes, allowing the dynamic tokens $d_i$ to progressively encode richer, frame-specific temporal details. While EncDiff applied similar ideas in static contexts, our work introduces and demonstrates its efficacy explicitly in sequential data.

\paragraph{3. Cross-Attention Interaction in U-Net}
Motivated by the success of cross-attention for semantic alignment in EncDiff, we significantly adapt this mechanism within our video-based diffusion decoder. Each U-Net denoising block incorporates structured cross-attention conditioned on the static token $s$ (global) and the frame-specific dynamic tokens $d_i$ (local). This selective routing ensures the static token consistently influences every frame, while dynamic tokens only affect their respective frames. Our novel formulation clearly separates and reinforces the distinct roles of static and dynamic factors.

Collectively, these carefully engineered biases significantly enhance disentanglement quality, surpassing prior methods.

\subsection{End-to-End Training}
\label{sec:training}

\ours{} is trained end-to-end in a single stage, optimizing the sequence encoder and the diffusion decoder simultaneously. Our objective function has two complementary components. First, we adopt the standard simplified DDPM loss widely used in diffusion models \cite{ho2020denoising}:
\begin{equation}
\mathcal{L}_{\text{simple}} = \mathbb{E}_{x_0\sim q(x_0),\,\epsilon_t\sim\mathcal{N}(0,I),\,t}\left[\|\epsilon_\theta(x_t,t,s,d_i)-\epsilon_t\|_1\right].
\end{equation}

Additionally, to further enforce orthogonality and explicitly discourage static-dynamic information leakage, we introduce a regularization term encouraging independence between the static token and each dynamic token:
\begin{equation}
\mathcal{L}_{\text{orth}} = \sum_{i=1}^{\nframes}(s^\top d_i)^2.
\end{equation}

Our final training objective combines both losses:
\begin{equation}
\mathcal{L} = \mathcal{L}_{\text{simple}} + \lambda\mathcal{L}_{\text{orth}},
\end{equation}
where the hyperparameter $\lambda$ is empirically determined to balance disentanglement quality and reconstruction fidelity. 

This combined training strategy ensures robust, end-to-end disentangled representation learning in our diffusion-based video framework.

\begin{figure*}[h]
	\centering
	\setlength\tabcolsep{6pt}         
	\renewcommand{\arraystretch}{0.9} 
	\begin{tabular}{l c}
		\textbf{Source}  &
		\includegraphics[width=0.85\textwidth]{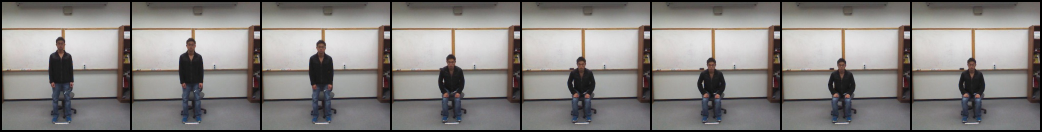} \\[6pt]
		\textbf{Target} &
		\includegraphics[width=0.85\textwidth]{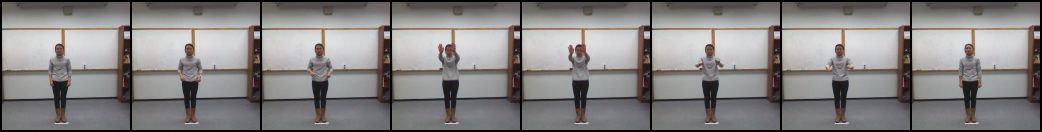} \\[6pt]
		\textbf{Ours}           &
		\includegraphics[width=0.85\textwidth]{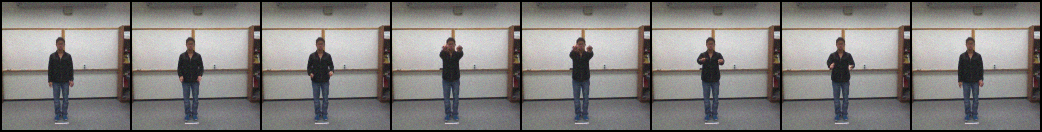} \\[6pt]
		\textbf{DBSE}           &
		\includegraphics[width=0.85\textwidth]{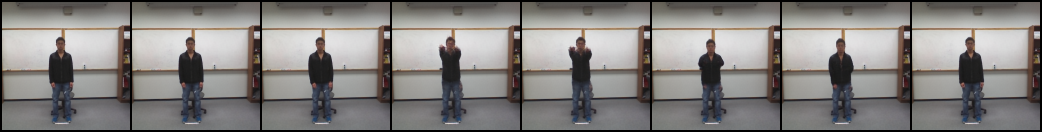} \\[6pt]
		\textbf{SPYL}           &
		\includegraphics[width=0.85\textwidth]{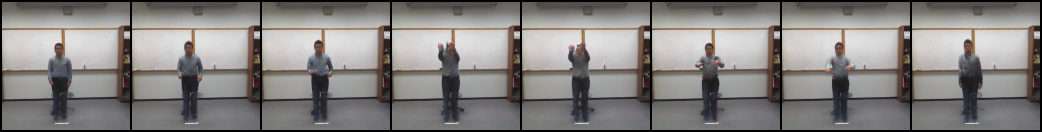} \\
	\end{tabular}	
	\caption{Static–dynamic factor swapping on a MHAD example. From top to bottom: static source (first frame repeated for our method and DBSE; full sequence for SPYL), dynamic source, and swapped outputs by our method, DBSE, and SPYL. Our method cleanly preserves identity while transferring the action; DBSE retains identity but fails to transfer motion; SPYL transfers motion at the cost of appearance fidelity.}
	\label{fig:swap_results}
\end{figure*}

\section{Experiments}
\label{sec:experiments}

\begin{figure*}[h]
	\centering
	\setlength\tabcolsep{6pt}         
	\renewcommand{\arraystretch}{0.9} 
	\begin{tabular}{l c}
		\textbf{Source}  &
		\includegraphics[width=0.85\textwidth]{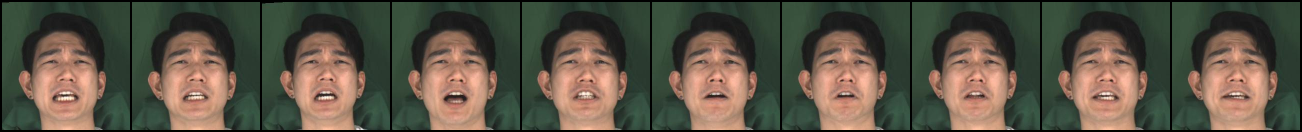} \\[6pt]
		\textbf{Target} &
		\includegraphics[width=0.85\textwidth]{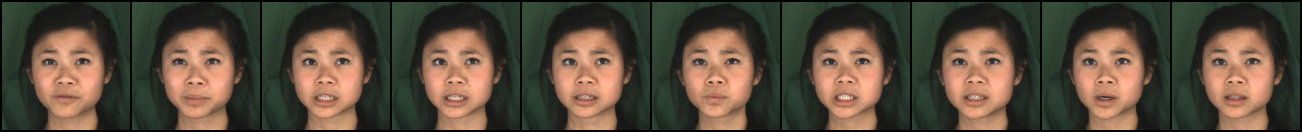} \\[6pt]
		\textbf{Ours}           &
		\includegraphics[width=0.85\textwidth]{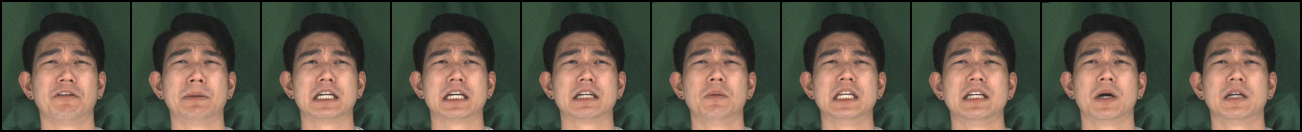} \\[6pt]
		\textbf{DBSE}           &
		\includegraphics[width=0.85\textwidth]{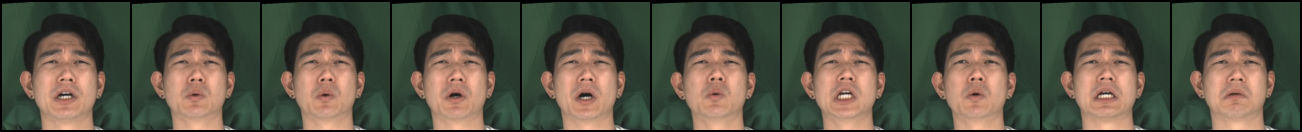} \\[6pt]
		\textbf{SPYL}           &
		\includegraphics[width=0.85\textwidth]{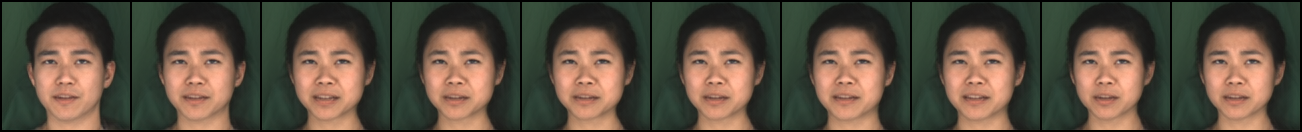} \\
	\end{tabular}
	\caption{Static–dynamic factor swapping on a MEAD example. From top to bottom: static source (first frame for our method and DBSE; full sequence for SPYL), dynamic source, and swapped outputs by our method, DBSE, and SPYL. Our method accurately transfers facial expressions while preserving identity; DBSE under-transfers expression dynamics; SPYL mixes appearance and motion, losing fidelity on both.}
	\label{fig:mead_swap_results}
\end{figure*}
\subsection{Experimental Setup}
\label{sec:setup}

\paragraph{Datasets.}
We evaluate our method on two standard real-world video datasets: the \textbf{MHAD}~\cite{chen2015utd} action-recognition dataset and the \textbf{MEAD}~\cite{wang2020mead} facial-expression dataset.

\textbf{MHAD} (UTD Multimodal Human Action Dataset) contains 861 video sequences captured with a Microsoft Kinect camera at a resolution of $640\times480$ pixels. The dataset includes 8 subjects (4 females, 4 males) performing 27 diverse actions such as waving, sitting, throwing, and walking. Each action is repeated four times by each subject. Following prior works, we standardize sequence lengths by randomly sampling 10-frame clips from the original sequences.

\textbf{MEAD} consists of image sequences featuring 30 subjects performing eight distinct facial expressions: anger, fear, disgust, happiness, sadness, surprise, contempt, and neutral. Video lengths vary across clips. Similar to MHAD, we standardize clips by randomly sampling 15 frames per video. Faces are then detected using Haar Cascades and cropped to isolate facial regions, finally resized to $128 \times 128$ pixels for processing.

\paragraph{Implementation Details}

Our model architecture consists of four main components: an image encoder, a static encoder, a dynamic encoder, and a conditional diffusion model.

\textbf{Image Encoder.}  
Each frame is independently encoded into a low-dimensional latent space using a convolutional encoder with three resolution levels and channel multipliers $(1, 2, 4)$. Each level includes two residual blocks, with a base channel width of 128. The encoder processes $128 \times 128$ RGB frames and outputs features projected to an embedding dimension of 3 via a $1 \times 1$ convolution.

\textbf{Static Encoder.}  
The static encoder extracts time-invariant content shared across the entire sequence. Following a frame-wise subtraction strategy, the latent representation of the first frame is subtracted from the other frames to isolate static content. The static code $s$ is derived by feeding the first frame’s embedding through a two-layer MLP with hidden dimension 1024 and ReLU activations, followed by a linear projection to produce a 256-dimensional feature.


\textbf{Dynamic Encoder.}  
The dynamic encoder models the time-varying content within the sequence. After removing static content by subtraction, the residual vectors are linearly projected to 256 dimensions, augmented with positional encodings (dropout \(0.1\)), and processed by a {Transformer encoder} (4 layers; \(d_{\mathrm{model}}=256\); \(n_{\text{heads}}=32\); feed-forward size \(1024\); dropout \(0.1\)). The encoder uses residual connections and layer normalization. A final linear head returns 256-dimensional frame-specific dynamic embeddings.

\textbf{Conditional Diffusion Model.}  
The denoising network is a UNet-based architecture that reconstructs clean video frames from their noisy versions through iterative denoising. The UNet includes four resolution levels with channel multipliers $(1, 2, 4, 4)$ and a base channel dimension of 128. Each level contains two residual blocks, and self-attention is applied at resolutions 32, 16, and 8. The network uses FiLM-like scale-shift normalization and residual blocks with learned up/downsampling.

Conditioning is achieved via cross-attention in spatial transformer blocks integrated at multiple UNet layers. These transformers take the concatenated static token $s$ (shared across all frames) and dynamic tokens $d_{1:\nframes}$ (frame-specific) as context. The tokens are projected to match the UNet’s attention dimension ($d=256$) and guide the denoising process. Time-step embeddings are encoded with a two-layer MLP and injected into residual blocks. The middle block, along with the downsampling and upsampling stages, integrates contextual attention, enabling adaptive use of both static and dynamic factors. The final output is a denoised frame with the same dimensions as the input.

\textbf{Hyperparameter Selection.}  
Prior works on static/dynamic sequential disentanglement often tune hyperparameters using full supervision, including labels for all factors, to find the best validation accuracy. This practice introduces bias, as it conflicts with the goal of unsupervised disentanglement and leaks label information into training. To avoid this issue, we deliberately avoid dataset-specific hyperparameter tuning. Instead, we select hyperparameters that work robustly across all datasets for both our method and baseline methods, ensuring a fair and label-agnostic evaluation.

\textbf{Baselines.}  
We compare \ours~against recent state-of-the-art methods, including SPYL~\cite{naiman2023sample} and DBSE~\cite{berman2024sequential}, both of which build on variational autoencoder (VAE) frameworks. In preliminary experiments, we found that the original implementations produced low-quality reconstructions and insufficient spatial detail, which could unfairly disadvantage the baselines. To ensure meaningful and fair comparisons, we adapted their architectures by replacing the original encoders and decoders with the same encoder-decoder backbone used in our method. Specifically, the encoder is identical to our image encoder, while the decoder mirrors its structure with upsampling blocks and integrated spatial attention at selected resolutions.

To further improve reconstruction quality, we augmented the standard pixel-wise reconstruction loss (typically MSE or $\ell_1$) with a perceptual loss computed from intermediate features of a pretrained VGG network. This encourages better preservation of semantic structure and fine-grained visual details. For all baselines, we use the same loss coefficient weights: 10 for the reconstruction loss, 5 for the static KL term, and 1 for the dynamic KL term.

\subsection{Results}
\label{sec:results}
%
%

\begin{table*}[h]
	\centering
	\begin{tabular}{lccc||c}
		\toprule
		\textbf{Model} & \textbf{Static Only (\%) $\uparrow$} & \textbf{Dynamic Only (\% $\uparrow$)} & \textbf{Joint Acc. (\%) $\uparrow$} & \textbf{Average Leakage\ (\%) $\downarrow$} \\
		\midrule
		DBSE   & 98.51 & 16.34 & 16.34 & 84.54 \\
		SPYL   & 40.59 & 45.54 & 16.83 & 99.47\\
		\ours   & 98.51 & 31.19 & 30.20 & 70.07\\
		\bottomrule
	\end{tabular}
	\caption{Swap‐based disentanglement on MHAD. Joint accuracy requires both identity and action to be correct; marginal accuracies measure each factor independently. Leakage is the average of identity‐into‐motion and motion‐into‐identity leakage rates.}
	\label{tab:swap_results}
	\vspace{-10pt}
\end{table*}

\subsubsection{Qualitative Results}

We qualitatively assess disentanglement by swapping dynamic factors between a \emph{source} sequence \(x_{1:\nframes}^\mathrm{src}\) (providing the static component) and a \emph{target} sequence \(x_{1:\nframes}^\mathrm{tgt}\) (providing the dynamic component). Concretely, we reconstruct frames conditioned on \((s^\mathrm{src},\,d_{1:\nframes}^\mathrm{tgt})\), holding the source appearance fixed while importing the target motion. Ideally, the resulting video retains the identity of the source actor performing the target action.

Figure~\ref{fig:swap_results} shows a single MHAD example. The top row displays the static source: for our method and DBSE this is simply the first frame repeated, whereas SPYL encodes the entire source clip. The next row shows the full target action. The three rows below present, in order, the outputs of our method, DBSE, and SPYL after swapping. Our approach retains the subject’s identity while seamlessly reproducing the target action. By contrast, DBSE preserves identity but fails to transfer the motion, and SPYL transfers motion at the expense of altering the actor’s appearance.

Notice that DBSE and SPYL both reconstruct the chair from the static source sequence—even though the chair is irrelevant to the target action. By contrast, our method focuses more on the dynamics: it omits the chair entirely, improving disentanglement by not carrying over background elements that do not pertain to the target dynamics.

Figure~\ref{fig:mead_swap_results} illustrates a similar swap on MEAD. Again, the first frame is used as the static source for ours and DBSE (with SPYL using the full sequence), and the second row shows the target expression sequence. The reconstructed outputs appear in the subsequent three rows for our method, DBSE, and SPYL. Our method faithfully preserves each subject’s identity and accurately transfers the target facial expression.
DBSE maintains identity but fails to reproduce the full dynamics of the expression dynamics, and SPYL fails on both counts—neither preserving the actor’s appearance nor accurately rendering the target expression, instead blending characteristics of source and target.

\subsubsection{Swap‐based disentanglement evaluation}
To quantify how well our model separates static appearance from dynamic motion, we perform a \emph{swap test} on held‐out MHAD clips. Given two clips \(x^1_{1:\nframes}\) and \(x^2_{1:\nframes}\) with ground‐truth subject identities \(s_1,s_2\) and action labels \(d^1,d^2\), we:

\begin{enumerate}
	\item Encode each clip into a static code \(s\in\mathbb{R}^{256}\) and dynamic codes \(\{d_i\}\in\mathbb{R}^{T\times256}\).
	\item Swap factors to synthesize
	\[
	\tilde x^{d_1s_2}_{1:\nframes} = \mathrm{Decode}(\{d^1_i\},\,s_2),\quad
	\tilde x^{d_2s_1}_{1:\nframes} = \mathrm{Decode}(\{d^2_i\},\,s_1).
	\]
	Ideally, \(\tilde x^{d_1s_2}\) should display the action of \(x^1\) (e.g.\ “right high wave”) in the identity of \(x^2\), and vice versa.
	\item Classify each \(\tilde x\) with a fixed, pre‐trained network to obtain predicted subject and action labels.
\end{enumerate}

We report three metrics over all swap pairs:
\begin{itemize}
	\item \textbf{Static‐only accuracy:} fraction where the predicted identity matches the swapped‐in subject.
	\item \textbf{Dynamic‐only accuracy:} fraction where the predicted action matches the driving action.
	\item \textbf{Joint accuracy:} fraction where both identity and action are correct.
\end{itemize}

%

As Table~\ref{tab:swap_results} shows, DiViD achieves the best joint swap accuracy, demonstrating true disentanglement of both static appearance and dynamic motion. While SPYL posts a high dynamic-only score, its very low static-only accuracy reveals that it simply reproduces the target motion at the expense of source identity—effectively ignoring the static factor. DBSE, on the other hand, perfectly preserves identity but fails to transfer any dynamics. DiViD strikes the optimal balance: it preserves source appearance and more than doubles DBSE’s dynamic accuracy, yielding the strongest joint performance overall.

Figure~\ref{fig:spyl_failure} further exposes SPYL’s failure mode: its “swapped” outputs are nothing more than direct copies of the target sequence, which artificially inflate its dynamic-only metric without any genuine factor separation. This clearly illustrates that dynamic-only accuracy can be highly misleading—real disentanglement requires strong performance on both factors, as captured by the joint accuracy.

\begin{figure}[h]
	\centering
	\setlength\tabcolsep{4pt}         
	\renewcommand{\arraystretch}{0.9} 
	\begin{tabular}{@{}l@{}}
		\textbf{Source}\\
		\vspace{-4pt}
		\includegraphics[width=\columnwidth]{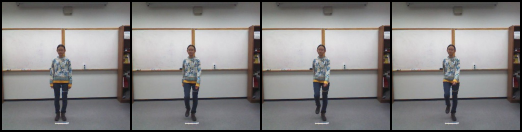}\\[4pt]
		\textbf{Target}\\
		\vspace{-4pt}
		\includegraphics[width=\columnwidth]{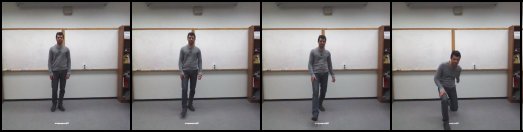}\\[4pt]
		\textbf{SPYL}\\
		\vspace{-4pt}
		\includegraphics[width=\columnwidth]{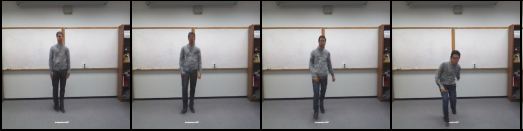}
	\end{tabular}
	\caption{Failure mode of SPYL on a MHAD example.  
		The top row shows the static source (first frame), the middle row the dynamic source, and the bottom row SPYL’s “swapped” output. Note how SPYL merely copies the target motion—failing to preserve the source identity—illustrating that high dynamic‐only accuracy can mask a lack of true disentanglement.}
	\label{fig:spyl_failure}
	
\end{figure}

\begin{table*}[h]
	\centering
	\begin{tabular}{lccc}
		\toprule
		\textbf{Method} & \textbf{Static Only (\%)} & \textbf{Dynamic Only (\%)} & \textbf{Joint Acc. (\%)} \\
		\midrule
		\ours~w/o Orth   & 100.0 & 13.4 & 13.4\\
		\ours~w/ LSTM    &  52.0 &  3.0 &  1.0\\
		\ours~w/ AdaGN   &  98.0 & 24.3 & 23.8\\
		\ours~w/linear   & 100.0 & 22.3 & 22.3\\
		\ours            &  98.5 & 31.2 & 30.2\\
		\bottomrule
	\end{tabular}
	\caption{Ablation study on MHAD. Joint accuracy requires both identity and action to be correct; marginal accuracies measure each factor independently. Leakage is the average of identity‐into‐motion and motion‐into‐identity leakage rates.}
	\label{tab:ablation}
\end{table*}

\subsubsection{Cross‐Leakage Classification}
To quantify how well each model disentangles static (subject identity) from dynamic (action) information, we freeze the encoder and train two lightweight classifiers.

\begin{itemize}
	\item \textbf{Static→Dynamic (S→D)}: predict \(y_{\mathrm{dynamic}}\) from \(s\) (measures action leakage into the static code).
	\item \textbf{Dynamic→Static (D→S)}: predict \(y_{\mathrm{static}}\) from \(\{d_i\}\) (measures identity leakage into the dynamic code).
\end{itemize}

From these two accuracies we compute the following metric:
\begin{equation*}
	\begin{split}
		\mathrm{Average\ leakage}  &= \tfrac12\bigl(\mathrm{Acc}_{S\to D}+\mathrm{Acc}_{D\to S}\bigr).
	\end{split}
\end{equation*}

Lower average leakage indicates minimal cross-leakage between static and dynamic codes.

This metric is reported in the last column of Table~\ref{tab:swap_results}.
SPYL shows very high leakage, indicating entangled representations. DBSE still leaks substantially. Our approach reducess average leakage by around 14 percentage points, demonstrating the most effective separation of static appearance and dynamic motion.

\subsection{Ablation Study}
\label{sec:abla}

We ablate the main design choices of \ours~on MHAD to quantify the contribution of the inductive biases (Sec.~\ref{sec:biases}) and the orthogonality regularizer introduced in our training objective (Sec.~\ref{sec:training}). Specifically, we vary (i) the orthogonality regularizer between static and dynamic tokens, (ii) the temporal encoder used for dynamic residuals, (iii) the conditioning pathway into the U-Net (cross-attention vs.\ AdaGN), and (iv) the variance ($\beta$) scheduler. Residual subtraction and the shared-noise setting are held fixed across all variants to isolate the effect of each change.\\

\noindent\textbf{Effect of orthogonality.}
Removing $\mathcal{L}_{\text{orth}}$ (\ours~w/o Orth) sharply reduces Dynamic-only ($13.4\%$ vs.\ $31.2\%$ for \ours) and Joint accuracy ($13.4\%$ vs.\ $30.2\%$). This confirms that $\mathcal{L}_{\text{orth}}$ is critical to prevent static information from leaking into the dynamic code.

\noindent\textbf{Temporal encoder.}
Replacing the Transformer encoder with a BiLSTM (\ours~w/ LSTM) collapses dynamic performance (Dynamic-only $3.0\%$, Joint $1.0\%$) under the same training setup, indicating that long-range temporal aggregation and rich token–token interactions from the Transformer are essential for our residual sequence.

\noindent\textbf{Conditioning pathway (U-Net).}
Substituting cross-attention with AdaGN (\ours~w/ AdaGN) degrades Dynamic-only ($24.3\%$) and Joint ($23.8\%$) relative to \ours. This supports our claim that cross-attention provides a stronger inductive bias for clean routing of global static $s$ vs.\ frame-specific $d_i$ than global feature modulation.

\noindent\textbf{Variance (\boldmath$\beta$) scheduler.}
We examine how the noise variance schedule shapes the time-varying bottleneck in our diffusion decoder. Because we inject the \emph{same} noise realization across frames, the $\beta$ schedule (and thus the SNR curve) controls how much temporal signal survives at each step, modulating the relative influence of $s$ vs.\ $d_i$ (see the discussion on schedule-dependent bottlenecks in diffusion). \ours\ with a cosine schedule (default) attains higher \emph{Dynamic-only} and \emph{Joint} accuracy than a linear schedule: from $22.3\%\!\to\!31.2\%$ (Dynamic) and $22.3\%\!\to\!30.2\%$ (Joint), while \emph{Static-only} remains near saturation ($\approx\!100\%$ vs.\ $98.5\%$). This indicates that a gentler mid-range SNR decay (cosine) gives cross-attention more opportunity to route motion through $d_i$, improving disentanglement, consistent with prior analyses that different $\beta$ schedules induce different information bottlenecks~\cite{yang2024diffusion}.\\

\noindent\textbf{Summary.}
\ours\ attains the best Joint accuracy ($30.2\%$) by combining: (i) residual subtraction to suppress static leakage at the source, (ii) a Transformer encoder for dynamic tokens, (iii) cross-attention conditioning in the U-Net, (iv) orthogonality regularization, and (v) an appropriate $\beta$ schedule that preserves a useful mid-SNR window for motion routing. Removing $\mathcal{L}_{\text{orth}}$, replacing the Transformer with LSTM/linear encoders, or substituting cross-attention with AdaGN consistently degrades Dynamic-only and Joint metrics under our swap-based protocol (Sec.~\ref{sec:results}).
\section{Conclusion}
\label{sec:conclusion}
We have presented \textbf{DiViD}, a novel end-to-end video diffusion framework for unsupervised static–dynamic disentanglement.  By combining a residual-based sequence encoder with a conditional DDPM decoder enriched by (i) a shared-noise schedule for temporal consistency, (ii) a time-varying KL bottleneck that naturally allocates capacity to static vs.\ dynamic factors, (iii) cross-attention to cleanly route global vs.\ frame-specific information, and (iv) an orthogonality regularizer, DiViD overcomes the information-leakage  of prior VAE-based approaches. Through the experiments on MHAD and MEAD, we have shown that DiViD achieves the highest joint swap accuracy, superior static fidelity, and reduced cross-leakage.
Future work will focus on (1) systematic ablations to quantify the impact of each inductive bias, (2) extending DiViD to conditional video synthesis, (3) integrating weak supervision signals to handle more complex, real-world datasets.

{
    \small
    \bibliographystyle{ieeenat_fullname}
    \bibliography{main}
}

\end{document}